\documentclass{article}
\usepackage{dcase2024,amsmath,graphicx,url,times,booktabs, tabularx}
\usepackage{multirow}
\title{From computation to consumption: exploring the compute-energy link for training and testing neural networks for SED systems}

\name{Constance Douwes, Romain Serizel}
\address{University of Lorraine, CNRS, Inria, Loria, 54000, Nancy, France\\constance.douwes@inria.fr, romain.serizel@loria.fr}
\begin{document}

\ninept
\maketitle

\begin{sloppy}

\begin{abstract}
The massive use of machine learning models, particularly neural networks, has raised serious concerns about their environmental impact. Indeed, over the last few years we have seen an explosion in the computing costs associated with training and deploying these systems. It is, therefore, crucial to understand their energy requirements in order to better integrate them into the evaluation of models, which has so far focused mainly on performance. In this paper, we study several neural network architectures that are key components of sound event detection systems, using an audio tagging task as an example. We measure the energy consumption for training and testing small to large architectures and establish complex relationships between the energy consumption, the number of floating-point operations, the number of parameters, and the GPU/memory utilization.
\end{abstract}


\section{Introduction}

Deep learning (DL) has become the principal focus of audio processing research, with numerous applications spanning various domains including sound event detection (SED) \cite{kim2023semi,berghi2024fusion}, speech recognition \cite{baevski2020wav2vec,radford2023robust} and music generation \cite{agostinelli2023musiclm,caillon2021rave}. As models become increasingly powerful and datasets grow larger, the associated computational costs have exploded \cite{amodei2018ai, sevilla2022compute,thompson2007computational}. Yet, the true cost of computation often remains obscured, as many computations are carried out on remote infrastructures or data centers. Nevertheless, these energy-intensive processes involved in training and deploying high-performance models have a real environmental footprint linked to their demand for electricity \cite{strubell2019energy, luccioni2023estimating}. This raises significant concerns in the current context of climate change and efforts to limit global warming to below 2 degrees \cite{parisagreement}. Even though models used in audio processing are smaller than those used in natural language processing, they still present similar problems \cite{douwes2023quality, parcollet2021energy}. 

The trends described above are driven by an ongoing pursuit of outperforming previous state-of-the-art systems, even by a small margin. Recently, there has been a slight shift towards reporting and quantifying the environmental costs associated with these advances \cite{henderson2020towards, schwartz2020green}. In the audio processing domain in particular, significant efforts have been made to balance performance and energy in the context of sound event detection \cite{serizel2023performance, ronchini2023performance} or speech recognition~\cite{parcollet2021energy}, and to emphasize the importance of considering quality metrics alongside energy footprint assessments in speech synthesis \cite{douwes2023quality}. All of these studies call for a fair and reliable metric to assess the computational footprint that reflects the energy consumption while being hardware independent to enable accurate comparisons between models. Although work such as Speckhard et al.~\cite{speckhard2022neural} shows a strong correlation between computational cost and energy consumption during inference for convolution-based models, to our knowledge similar investigations have not been conducted for training or for other architectures. Even if a few hundred experiments are sometimes required to train a model, the cost of the training phase represents only 10\% to 20\% of the total CO2 emissions of the associated machine learning usage, with the majority occurring during the inference phase \cite{wu2022sustainable}. However, as audio processing researchers, the majority of our energy consumption lies in the training phase, and should not be overshadowed. 

In this article, we aim to understand the computational factors that impact the energy consumption for the training or testing deep learning models that compose SED systems. This study is conducted in the context of the DCASE challenge task 4, where participants have been required since 2022 \cite{ronchini2022description} to report their energy consumption alongside computational factors such as the number of parameters and the number of operations. Specifically, we seek an indicator that can estimate the energy consumption based on computational measurements. This would allow us to estimate each system's consumption on the same hardware and provide fair comparisons between systems, extending the work of Ronchini et al.\cite{ronchini2023performance}. We focus our analysis on well-known architectures such as MLP, RNN, CNN and CRNN. CRNN is specifically the current architecture used in Task 4 of the DCASE Challenge \cite{dcase2023Task4a}. We compute the number of parameters of the models and the number of floating point operations (FLOPs) as two potential candidate factors for energy consumption estimation. We show that as the number of operations increases, so does the energy consumption across all architectures during both the test and training phases. However, the relative increase in energy consumption varies between architectures and phases. We identify two distinct trends: one for MLP/RNN, and one for CNN/CRNN. Finally, we identify relationship between energy consumption and GPU utilization during both training and testing phases, which could serve as a basis for future research on computational metrics.

In summary, our key contributions are :
\begin{itemize}
    \item A comparative analysis of prominent architectures (MLP, CNN, RNN, CRNN) and their associated energy consumption.
    \item The identification of two distinct trends in energy consumption based on architecture type, notably distinguishing between MLP/RNN and CNN/CRNN architectures.
    % \item An strong correlation between FLOPs, number of parameters and energy for MLP/RNN.
    \item A relative comparison of power usage between training and test stages.
\end{itemize}

\section{Methodology}

Computing and monitoring the computational and energy costs of the two phases of deep learning systems - training and inference - is a complex endeavour. We present here our methodology for assessing both, mentioning previous work in these areas.
% and the justification for the metrics used in the paper.

\subsection{Computational cost}
Traditional methods rely on metrics such as the size of the model (the number of parameters) and the number of floating-point operations (FLOPs) computed by the model to estimate the computational cost. While computing the number of parameters (or weights) of a model is straightforward, computing the number of operations can be a difficult task, especially for complex architectures, and this number is very sensitive to the size of the input/output. At inference, only forward calculations are performed, so the number of operations is the sum of all operations across all layers. We use the deepspeed profiler ~\cite{rasley2020deepspeed} to quantify these forward pass operations accurately. In contrast, training is a more complex process involving iterative forward and backward calculations. In particular, the backward pass also computes the gradient with respect to the parameters, the loss and update the weights. However, at the time of writing, no profiler provided the exact number of backward operations, so we derive this number using the ratio 2:1 as an approximation~\cite{epoch2021backwardforwardFLOPratio}. In total, the number of operations of a training iteration (forward and backward) is three times the number of operations of an inference (forward only). 
%In this paper we focus on the number of floating-point operations, although an alternative metric is the count of multiply-accumulate operations (MAC), which combines both multiplication and addition operations. However, there is a consensus among profilers that 1 MAC is equivalent to 2 FLOPs, which justifies our choice of only taking FLOPs into account.

\subsection{Energy consumption}
Several Python trackers have emerged to facilitate the computation of energy consumption~\cite{jay2023experimental}. In most of the trackers, the total consumption is calculated as the sum of the consumption of each component of the computer: GPU, CPU and RAM. In our study, we focus specifically on analysing the energy consumption of the GPU given by CodeCarbon ~\cite{schmidt2021codecarbon}. Indeed, preliminary experiments have led us to conclude that while GPU power fluctuates, CPU power remains stable. Regarding ram energy, CodeCarbon estimates 3 watts per 8 GB, which also remains constant over time. We made sure that any increases in GPU power with the python trackers were correlated with energy consumption monitored on the system's baseboard management controller (BMC). We also monitor the GPU and memory utilization from Nvidia SMI query every 5 seconds to get the mean uses of the each experiment. 

\section{Experiments}

Our objective is to better understand the energy consumption at train and test and to relate it to computational cost of a given model and architecture. To achieve this, we evaluate different types and sizes of architectures for audio tagging systems.\footnote{\url{https://github.com/ConstanceDws/toolbox_energy}}

 % In the following, we present the dataset, followed by the descriptions of the models and architectures used, and details of the training and inference procedures

% \subsection{Dataset}
% We used the Speech Commands Zero through Nine (SC09) dataset from \cite{goel2022s}, which is a subset of the larger Speech Commands dataset \cite{warden2018speech}. It is composed of 1-second audio clips of spoken digits from zero to nine pronounced by different speakers under different noise conditions and accents. Each class (digit) contains about 4000 files, resulting in a total duration of about 11 hours. The audio clips are sampled at 16kHz. In our experiments, we use the mel-spectrogram representation, which is the most commonly used representation in speech processing as it provides a compact representation of the spectral content of audio signals. We compute the mel-spectrogram using 128 bands, with an FFT size of 2048 and a hop size of 256. We use the first 64 frames as the input of the digit classifier.
\subsection{Task description} 
Audio tagging involves assigning one or multiple tags to an audio signal without any temporal information. For this experiment, we work on the real part of the DESED dataset \cite{turpault2019sound}. This dataset contains 10-second audio clips recorded in domestic environments. We convert those recordings into mel-spectrogram representations with 128 bands, an FFT size of 2048 and a hop size of 256. We only take the first 64 frames as input, which corresponds to approximately the first 1 second of the audio signal. Although this significantly impacts the performance of the model, it reduce the system's complexity, allowing for more lightweight experiments, as we do not focus on performance but only on energy.

\subsection{Models} We implement four neural network architectures: multi-layer perceptron (MLP), convolutional neural network (CNN), recurrent neural network (RNN), and convolutional recurrent neural network (CRNN). For the MLP, we implement a series of linear layers followed by ReLU activation functions. For the CNN, we adopt a sequence of Conv2d, ReLU and MaxPool2d layers. For the RNN we use GRU cells and for the CRNN we start with Conv2d, ReLU and MaxPool2d layers followed by a GRU cell. All implementations are completed with a final linear layer and a sigmoid activation function that outputs a probability vector for the 10 classes. For each architecture, we systematically increase the number of layers and adjust the hidden sizes per layer, gradually scaling up to reach the full GPU memory capacity and utilization, resulting in 43 models. We present the summary of all the configurations tested in Table \ref{tab:model}. We intentionally chose those configurations to achieve meaningful variations in the number of FLOPs without conducting redundant experiments.

\begin{table}[t]
    \centering
    \begin{tabular}{ccc}
        %\hline
        \textbf{Model} & \textbf{Num Layers} & \textbf{Hidden Sizes} \\
        \hline
        \multirow{3}{*}{MLP} & 1 & 512, 1024, 2048 \\
        & 4 & 1024, 2048, 4096 \\
        & 6, 10, 16, 32 & 4096 \\
        \hline
        \multirow{3}{*}{CNN} & 1 & 128, 256, 512, 1024 \\
        & 2 & 128, 256, 384, 512, 768, 1024 \\
        & 6 & 384, 768 \\
        \hline
        \multirow{3}{*}{RNN} & 1 & 128, 512, 1024, 2048 \\
        & 4, 6& 1024, 2048 \\
        & 2, 10, 14 & 2048 \\
       \hline
        \multirow{3}{*}{CRNN} & [1,1], [2,1], [1,2] & [64,64], [256,64], [512, 256]  \\
        & [2,2] & [728, 256]\\
        & [1,2], [2,2] & [1024, 256] \\
    \end{tabular}
    \caption{Summary of all the configurations tested in our experiment. For each number of layer, we tested different hidden sizes. For CRNN, the configurations first indicate the convolutional layers and then the recurrent layers.}
    \label{tab:model}
\end{table}

\subsection{Training and test}

Our experiments diverge from the conventional research of accuracy performance. Instead, we train all models for a single epoch on the same Nvidia Tesla T4 GPU and monitor the energy of the training phase. To focus solely on architectural differences, we use a consistent batch size of 8. Although the choice of criterion, optimizer, and learning rate is crucial for model convergence, it does significantly impact energy measurements. Therefore, we employ the cross-entropy function as the criterion, fix the learning rate at $10^{-3}$, and use the ADAM optimizer \cite{kingma2014adam}. We did not include any validation steps in the training routine to isolate the effects of training. Instead, we measure the energy consumption during the test phase separately. The test phase involves running the model (inference) and computing the error. Although inference for such small models can generally be performed on the CPU, we ensure consistency with the training phase measurements by also running the test phase on the same Nvidia T4 GPU for the entire dataset (corresponding to 1 epochs of training).

\section{Results}

In this section, we explore the relationship between computational metrics and the energy consumption. Our analysis aims to identify trends and discrepancies in energy consumption at train and test across various architectures and configurations.

\subsection{Relationship between energy and computational cost at test}

\begin{figure}[t]
    \centering
    \includegraphics[width=\linewidth]{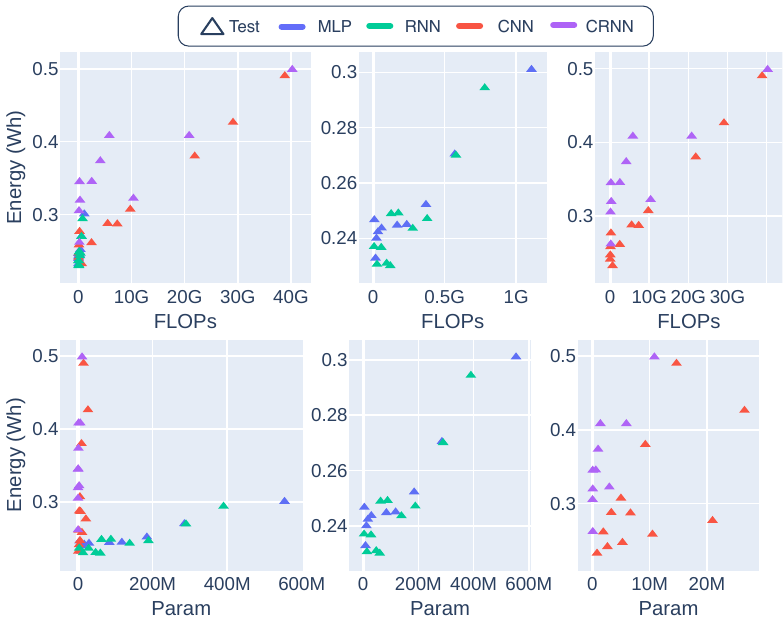}
    \caption{Energy consumption at test for various neural network architectures and configurations, as a function of FLOPs (top) and of parameters (bottom). The three columns show: (1) all architectures together, (2) only MLP/RNN (in blue and green), and (3) only CNN/CRNN (in red and purple).}
    \label{fig:energy_test}
\end{figure}

First, we examine the energy consumption of the test, as existing research suggests that there is a correlation between FLOPs and energy consumption for convolutional models \cite{speckhard2022neural} on CPU. Figure~\ref{fig:energy_test} shows the result of this experiment, where the top row presents the GPU energy consumption as a function of FLOPs, and the bottom row the energy consumption as a function of the number of parameters. The first row shows that increasing the number of operations at test leads to an increase in energy consumption for all types of architecture. A closer examination of each architecture type reveals that the relationship between FLOPs and energy consumption exhibits some affine patterns. Examining the number of parameters in the second row, significant disparities emerge between MLP/RNN and CNN/CRNN models: the relationship between the number of parameters and the energy consumption is almost affine for MLP/RNN (and similar to the relationship with FLOPs), but for CNN and CRNN the relationship is more chaotic. This discrepancy is mainly due to the architectural elements composing these networks. Convolutional layers use parameter sharing, which contrasts with fully connected layers where each parameter is unique to its connection. Similarly, in recurrent layers, the connections between units often have unique weights, although some forms of parameter sharing can occur as well. Consequently, MLP and RNN have a higher number of parameters but a lower number of operations relative to CNN. These observations suggest that the number of operations and the number of parameters are not reliable indicators for estimating energy consumption at test, regardless of the model type, as the affine patterns are not consistent across architectures. However, they could be useful within a single architecture scenario comparisons.

\subsection{Relationship between energy and computational cost at training}

\begin{figure}[t]
    \centering
    \includegraphics[width=\linewidth]{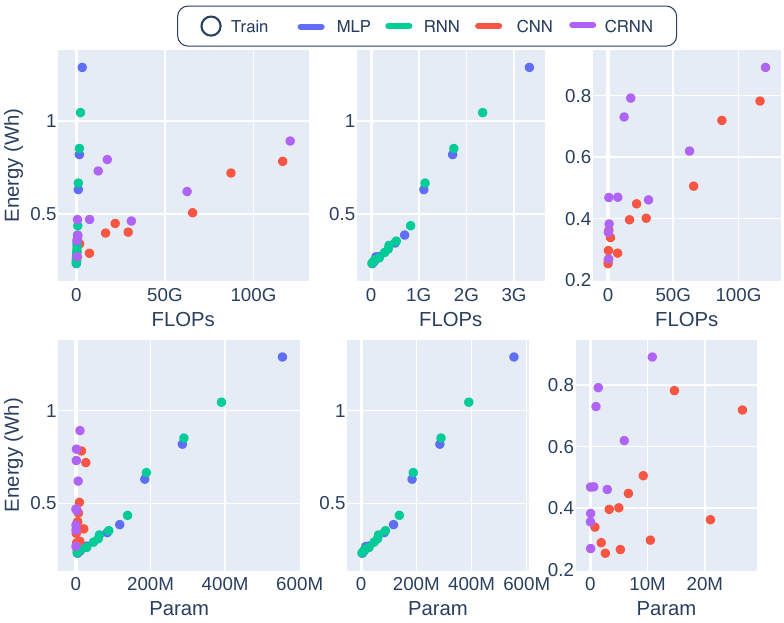}
    \caption{Energy consumption for training various neural network architectures and configurations, as a function of FLOPs (top) and of parameters (bottom). The three columns show: (1) all architectures together, (2) only MLP/RNN (in blue and green), and (3) only CNN/CRNN (in red and purple).}
    \label{fig:energy_train_flops}
\end{figure}

Building on our previous results, we now investigate the energy consumption associated with training. Figure \ref{fig:energy_train_flops} displays the energy consumption for training in function of the two computational metrics arranged as previously described. Regarding the interaction between energy and FLOPs, we observe two distinct trends. For MLP/RNN, the data points follow a steep curve on the left side, while for CNN, the curve smoothly increases and spans the entire plot. The CRNN architecture appears to exhibit characteristics that lie between the two aforementioned trends. In some configurations, the CRNN behaves as a CNN at higher FLOPs and as an RNN at lower FLOPs. A plausible explanation of this two trends could be the higher memory exchanges associated with MLP/RNN compared to CNN architectures that would cause higher energy consumption but do not increase the FLOPs. An important result is the almost affine relationship between FLOPs and energy consumption for MLP and RNN, suggesting that GPUs handle these architectures similarly during training causing close energy consumption for the same FLOPs. However, for CNN and CRNN, FLOPs alone do not provide a conclusive estimate of the energy consumption. Regarding the number of parameters, we conclude consistent results as for the test relationship. As a result, for the training consumption, neither FLOPs nor parameters are good estimators of energy consumption without specific knowledge of the model architecture, and one hypothesis could comes from the difference between the architectural elements of the network.

\subsection{Training and test comparisons}

\begin{figure}[t]
    \centering
    \includegraphics[width=\linewidth]{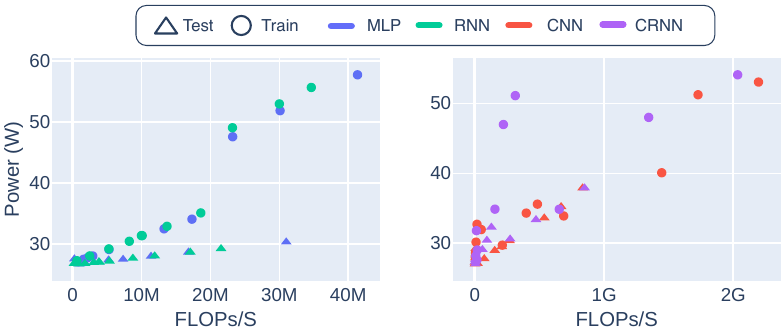}
    \caption{Average power during training (circles) and test (triangles) as a function of FLOPs/S.}
    \label{fig:train_test_power}
\end{figure}

% \begin{figure}[t]
%     \centering
%     \includegraphics[width=\linewidth]{plot/energy_flops.pdf}
%     \caption{Relationship between the average power and the GPU utilization at training and test.}
%     \label{fig:energy_train_2D}
% \end{figure}

To further investigate the link between energy and computation, we investigate the mean average power at test and train and relate it to the number of floating points operations per seconds. The average is calculated as the energy divided by the length of the experiment. We present the result of this analysis in Figure \ref{fig:train_test_power}, where the FLOPs/S is computed as the FLOPs divided by the duration of one epoch for training and test. We see that there is a nearly-affine relationship between FLOPs/S and power at test for the MLP/RNN architectures, as indicated by the aligned triangles. However, this affine relationship is less evident for training, as highlighted by a bend around 20M FLOPs/S. An significant result of this analysis is the disparity in average power consumption between MLP/RNN at train and test: circles are positioned higher on the plot, while triangles are lower and there is no overlap between the two sets. In contrast, for CNN and CRNN, triangles and circles occupy similar regions, indicating that MLP and RNN architectures require much more power for training than for testing compared to CNN/CRNN.

\subsection{GPU and memory utilization}

\begin{figure}[t]
    \centering
    \includegraphics[width=\linewidth]{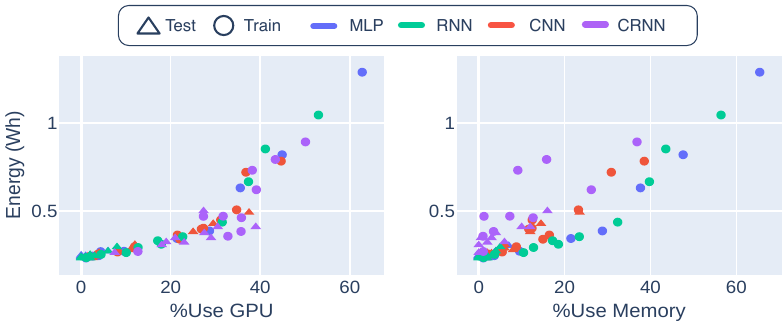}
    \caption{Relationship between the energy consumption and the GPU utilization (left) and memory utilization (right) for training and test.}
    \label{fig:use_gpu_mem}
\end{figure}

During our experiments, we also monitored the GPU and memory utilization given by Nvidia SMI. Figure \ref{fig:use_gpu_mem} illustrates the relationship between the energy and the GPU and memory utilization during both training and test phases. Notably, a strong correlation exists between GPU use and energy. What is noteworthy is that this correlation remains independent of the phase (train or test) and the architectures. This results in a metric that is highly recommended for estimating the energy consumption of a given model, although it is dependent on the hardware. It would be interesting to find a combination of the number FLOPs and the number of parameters that could reflect the GPU utilization. For memory utilization, the correlation is not as straightforward, but it shows that memory also has an impact on energy consumption, with a higher dependency on the architecture type than GPU utilization.

\section{Discussion and future works}

In this article, we specifically study the audio tagging task, using very simple architectures that are far from current SED models. It would therefore be interesting to explore more advanced models in the field and assess whether similar trends persist. In addition, the training procedure implemented here is one of the most conventional methods of deep learning, but recent advances have introduced much more complex procedures, resulting in higher computational costs and potentially different energy consumption. For example, using techniques such as teacher-student learning (used in the baseline) can lead to higher computational costs and therefore a different energy footprint. It is also important to note that energy consumption throughout our study is measured for a single epoch, and is therefore relative to the dataset. Experiments to determine whether there is a linear relation between data size and energy consumption would be recommended to remove the dependency on the dataset. 

Additionally, we focused here on a single hardware (one Nvidia Tesla T4). However, analyzing the differences within a single hardware configuration and exploring the variations between different hardware configurations could provide some additional information on the energy consumption. This approach could also contribute to efforts to normalize hardware energy measurements, such as those proposed by Serizel et al.~\cite{serizel2023performance}. Furthermore, our study did not address the performance of the models. It's likely that a CNN and CRNN may have different performances compared to an MLP or an RNN. This concept aligns with Douwes et al. \cite{douwes2021energy}, emphasizing the need to explore multi-objective criteria by considering factors such as model performance, energy consumption, and computational efficiency simultaneously.

\section{Conclusions}

Our study provides a better understanding of the relationship between computational cost and energy consumption for various neural networks used in SED tasks. We observed that while the number of floating-point operations and the number of parameters influenced energy consumption, these metrics were not consistent predictors across all architectures. We identify distinct trends and discrepancies in energy consumption during both testing and training phases, with notable differences between MLP/RNN and CNN/CRNN models. Finally, we establish correlations between energy consumption and GPU utilization for both training and test phases, that could lay as a foundation for future research on computational indicators.
We hope that this study will contribute to the development of green AI practices not only in speech processing but also across other domains.

% Either list references using the bibliography style file IEEEtran.bst
\bibliographystyle{IEEEtran}
\bibliography{refs}

\end{sloppy}
\end{document}